\documentclass[journal,transmag]{IEEEtran}
\usepackage{graphicx}
\usepackage{amsfonts}

\begin{document}
\title{Innovative Second-Generation Wavelets Construction With Recurrent Neural Networks for Solar Radiation Forecasting}

\author{\IEEEauthorblockN{Giacomo Capizzi\IEEEauthorrefmark{1},
Christian Napoli\IEEEauthorrefmark{2,*}, and
Francesco Bonanno\IEEEauthorrefmark{1}}
\IEEEauthorblockA{\IEEEauthorrefmark{1}Dpt. of Electrical, Electronic and Informatic Engineering, University of Catania, Italy}
\IEEEauthorblockA{\IEEEauthorrefmark{2}Dpt. of Physics and Astronomy, University of Catania, Italy}% <-this % stops an unwanted space
\thanks{*Email: napoli@dmi.unict.it.}
\thanks{PUBLISHED ON: \bf IEEE Transactions on Neural Networks and Learning Systems, vol. 23, N. 11, pp. 1805-1815, 2012}}

\markboth{Innovative $2^{nd}$ Generation Wavelet Construction with RNNs for Solar Radiation Forecasting -- PREPRINT}%
{Shell \MakeLowercase{\textit{et al.}}: Bare Demo of IEEEtran.cls for Journals}

 \begin{titlepage}
 \begin{center}
 {\Large \sc PREPRINT VERSION\\}
  \vspace{15mm}
 {\Huge Innovative Second-Generation Wavelets Construction With Recurrent Neural Networks for Solar Radiation Forecasting\\}
 \vspace{10mm}
 {\Large G. Capizzi, C. Napoli, and F. Bonanno\\}
 \vspace{15mm}
 {\Large \sc PUBLISHED ON: \bf IEEE Transactions on Neural Networks and Learning Systems, vol. 23, N. 11, pp. 1805-1815, 2012\\}
 \end{center}
 \vspace{10mm}
 {\Large \sc BIBITEX: \\}
 \vspace{5mm}
 
 @article\{Capizzi6320656, \\
author=\{Capizzi, G. and Napoli, C. and Bonanno, F.\}, \\
journal=\{Neural Networks and Learning Systems, IEEE Transactions on\}, \\
title=\{Innovative Second-Generation Wavelets Construction With Recurrent Neural Networks for Solar Radiation Forecasting\}, \\
year=\{2012\}, \\
volume=\{23\}, \\
number=\{11\}, \\
pages=\{1805-1815\}, \\
doi=\{10.1109/TNNLS.2012.2216546\}, \\
ISSN=\{2162-237X\},\\
\}
 \vspace{35mm}
 \begin{center}
Published version copyright \copyright~2012 IEEE \\
\vspace{5mm}
UPLOADED UNDER SELF-ARCHIVING POLICIES\\
NO COPYRIGHT INFRINGEMENT INTENDED \\
 \end{center}
\end{titlepage}

\IEEEtitleabstractindextext{
\begin{abstract}
Solar radiation prediction is an important challenge for the electrical engineer because it is used to estimate the power developed by commercial photovoltaic modules. This paper deals with the problem of solar radiation prediction based on observed meteorological data. A 2-day forecast is obtained by using novel wavelet recurrent neural networks (WRNNs). In fact, these WRNNS are used to exploit the correlation between solar radiation and timescale-related variations of wind speed, humidity, and temperature. The input to the selected WRNN is provided by timescale-related bands of wavelet coefficients obtained from meteorological time series. The experimental setup available at the University of Catania, Italy, provided this information. The novelty of this approach is that the proposed WRNN performs the prediction in the wavelet domain and, in addition, also performs the inverse wavelet transform, giving the predicted signal as output. The obtained simulation results show a very low root-mean-square error compared to the results of the solar radiation prediction approaches obtained by hybrid neural networks reported in the recent literature.
\end{abstract}

}

\maketitle

% –

\section{Introduction}
\IEEEPARstart{D}{aily} total solar radiation is considered as the most important parameter in meteorology, solar conversion, and renewable energy applications, particularly for the sizing of stand-alone photovoltaic (PV) systems \cite{r1}, \cite{r2}, \cite{r3}, \cite{r4}. The behavior of solar radiation is complex, either periodic or random, and the wavelet-transformed frequency components corresponding to various time–frequency domains of solar radiation show a similar behavior. This type of data is usually presented as a time series, whose prediction is an important scientific task. Situations where an underlying model generating the observed data is not known are especially challenging. Modeling a time series includes the stochastic prediction and the optimal prediction of a signal sample (in a minimum mean-square sense), given a finite number of past samples. In the literature, several methods to predict solar radiation have been reported, in particular statistical methods. The conventional statistical models can be considered as times-series-based models. These include auto-regressive (AR) and AR integrated moving average (ARIMA) models, Markov chains, and the Markov transitions matrix (MTM) approach. It is well known that these models are based on simplifying statistical assumptions about the measured data, which are not always true. Many researchers have attempted modeling solar radiation. The existing models established by classical approaches include, e.g., the so-called clear-day solar radiation, half-sine, Colares–Pereira and Rabl, and ARIMA hour-by-hour solar irradiation models \cite{r5}, \cite{r6}, \cite{r7}, \cite{r8}, \cite{r9}. Most existing models give relatively large errors and are sometimes difficult to use widely. Many authors have made important contributions to prove that recurrent neural networks (RNNs) are a powerful tool to exploit the statistical properties of time series \cite{r10},\cite{r11}, \cite{r12}.

In this paper, the use of the wavelet RNN (WRNN) architecture to find a forecasting model for the prediction of solar radiation is investigated. The total solar radiation is the most important parameter in the performance prediction of renewable energy systems, particularly in sizing PV power systems. Experimental setups for the measurement of solar radiation of assembled PV modules are available in the research laboratories of the University of Catania, Italy. The main objective of this paper is to investigate the WRNN architecture for modeling and prediction of the daily total solar radiation. In this paper, the improvement in the accuracy of a forecasting model is achieved by a wavelet-based transform. First, we decompose the sample data sequence of solar radiation into several components of various time–frequency domains according to wavelet analysis; then, we use the RNNs particularly established to make forecasts for all domains based on these components; and, finally, we arrive at the algebraic sum of the forecasts. Thereby, a relatively accurate forecast of solar irradiation could be achieved. Thus, by means of a combination of artificial neural networks (ANNs) with wavelet analysis, we arrive at a model to forecast solar radiation.

The rest of this paper is structured as follows. In Section II, we summarize the problem of solar radiation forecasting. Section III reports the basic elements of wavelet theory RNN architectures. Section IV presents the proposed novel structure for the prediction of solar radiation. The experimental setup is described in Section V. The simulation results are presented in Section VI. Finally, in Section VII, conclusions are drawn.

In this paper, the implemented WRNN is able to reconstruct a signal from wavelet coefficients; but if it is possible to predict these wavelet coefficients and then to reconstruct and predict the signal, then a better prediction can be obtained. The second-generation wavelets, which play an important role in the prediction problem, constitute a powerful tool to obtain a better prediction of solar radiation. Wavelet networks are feedforward networks using wavelets as activation functions. Such networks have been used successfully in various engineering applications such as classification, identification, and control problems \cite{r13}, \cite{r14}, \cite{r15}, \cite{r16}, and we present a novel contribution in this paper.

\section{Problem of Solar Radiation Forecasting}
The problem of renewable energy forecasting such as that of wind and wind power has been dealt with in recent literature, and very interesting contributions have been reported \cite{r17}, \cite{r18}, \cite{r19}. A number of different architectures and learning methods have been applied with various degrees of success to the problem of time-series prediction for total solar radiation. Elizondo et al.proposed the use of a feedforward neural network to estimate the daily solar radiation, the predicted daily total solar radiation, and the tilt angle versus panel efficiency of PV arrays, as well as other parameters such as temperature, precipitation, clear-sky radiation, day length, and day of the year \cite{r12}. Al-Alawi and Al-Hinai \cite{r9} have used an ANN to analyze the relationship between total radiation and climatology variables, and their model predicted total radiation values to a good accuracy of approximately 93\%. Guessoum et al. \cite{r20} used radial basis function (RBF) networks to predict solar radiation data. In their research, the input and the output are the solar radiation data corresponding to a particular day and those of the next day . The RBF model predicts solar radiation data using Kolmokorov–Simernov statistics to an accuracy of 1.36\%. Kemmoku et al. \cite{r10} used single and multistage neural networks to forecast the daily insolation, showing the mean error reduced from about 30\% (by the single stage) to about 20\% (by the 708 multistage). The input values to the network were latitude, longitude, altitude, and sunshine duration, and the results for the testing stations obtained were within 16\%. Kalogirou et al. \cite{r11} used an RNN to predict the maximum solar radiation from the relative humidity (RH) and temperature. The results indicated that the correlation coefficient obtained varied between 98.58\% and 98.75\%. Single-layer feedforward neural networks with hidden nodes of adaptive wavelet functions have been successfully demonstrated to have the potential for use in different applications. The combination of wavelet theory and neural networks has led to the development of wavelet networks \cite{r15}.

\section{Basics of Wavelet Theory and RNN Architectures}
This section is devoted to a brief review of the theory of wavelet analysis and RNNs. More detailed treatments can be found in \cite{r16}, \cite{r21}, and \cite{r22}.

\subsection{Wavelet Theory}
A multiresolution analysis of  is defined as a set of closed subspaces  with  that exhibit the following properties:

\begin{enumerate}
\item $V_j \subset V_{j+1}$
\item $v(x) \in V_j \Leftrightarrow v(2x) \in V_{j+1}$
\item $v(x) \in V_0 \Leftrightarrow v(x+1) \in V_0$
\item $\bigcup\limits_{j=-\infty}^{+\infty} V_j ~~ \mbox{is dense in}~ L^2(\mathbb{R})$
\item $\bigcap\limits_{j=-\infty}^{+\infty} V_j = \{\mathbf{0}\}$
\item $\exists~ \varphi(x) \in V_0 : \{ \varphi(x-l) | l \in \mathbb{Z}\}~~\mbox{is a Riesz basis of}~V_0$
\end{enumerate}

where $\varphi(x)$ is called scaling function. As a result of such properties, a sequence $\{h_k\}$ exists such that the scaling function satisfies a refinement equation

\begin{equation}
\varphi(x)=2\sum\limits_k h_k \varphi(2x-l)
\label{e1}
\end{equation}

The set of functions $\{\varphi_{j,l}(x)|l\in\mathbb{Z}\}$ with

\begin{equation}
\varphi_{j,l}(x)=\sqrt{2^{j}}\varphi (2^{j}\,x-l)
\end{equation}
is a Riesz basis of $V_j$. Define now $W_j$ as a complementary space of $V_j$ in $V_{j+1}$ , such that $V_{j+1}=V_j \oplus W_j$, $v(2x) \in W_{j+1}$, and $v(x) \in W_0 \Leftrightarrow v(x+1) \in W_0$. Consequently

\begin{equation}
\bigoplus\limits_{j=-\infty}^{+\infty} W_j = L^2(\mathbb{R})
\end{equation}

A function $\phi(x)$ is a wavelet if the set of functions $\{\varphi(x-l)|l\in\mathbb{Z}\}$ is a Riesz basis of $W_0$ and also meets the following two conditions:
\begin{equation}
\int\limits_{-\infty}^{+\infty}\psi (x) dx=0
\label{e2}
\end{equation}
and
\begin{equation}
\Vert\psi (x)\Vert^{2}=\int\limits_{-\infty}^{+\infty}\psi(x)\psi^{\ast}(x) dx=1
\label{e3}
\end{equation}

If the wavelet is also an element of $V_0$, a sequence $\{g_k\}$ exists such that
\begin{equation}
\psi (x)=2\sum_{k}g_{k}\,\varphi (2x-l)
\label{e4}
\end{equation}
The set of functions $\{\varphi_{j,l}(x)|l\in\mathbb{Z}\}$  is now a Riesz basis of $L^2(\mathbb{R})$. The coefficients in the expansion of a function in the wavelet basis are given by the inner product with dual wavelet 
$\tilde{\psi}_{j,l}(x)=\sqrt{2^j}\tilde{\psi}(2^jx-l)$ such that
\begin{equation}
f(x)=\sum_{j,l}\langle f,\widetilde{\psi}_{j,l}\rangle\,\psi_{j,l}(x)
\label{e5}
\end{equation}
Likewise, a projection on $V_j$ is given by

\begin{equation}
P_{j}f(x)=\sum_{l}\langle f,\widetilde{\varphi}_{j,l}\rangle\,\varphi_{j,l}(x)
\end{equation}

where $\tilde{\varphi}_{j,l}(x)=\sqrt{2^j}\tilde{\psi}(2^jx-l)$ are the dual scaling functions. The dual functions have to satisfy the biorthogonality conditions
\begin{equation}
\langle\varphi_{j,l},\widetilde{\varphi}_{j,l^{\prime}}\rangle=\delta_{l-l^{\prime}}\end{equation}
and
\begin{equation}
\langle\psi_{j,l},\widetilde{\psi}_{j^{\prime},l^{\prime}}\rangle=\delta_{j-j^{\prime}}\delta_{l-l^{\prime}}\end{equation}

They satisfy refinement relations similar to (\ref{e1}) and (\ref{e4}) involving sequences $\{\tilde{h}_k\}$ and $\{\tilde{g}_k\}$. In case the basis functions coincide with their duals, the basis is orthogonal.

\subsection{RNN Architectures}
For deterministic dynamical behaviors, the observation at a current time point can be modeled as a function of a certain number of preceding observations. In such cases, the model used should have some internal memory to store and update context information \cite{r23}, \cite{r24}. This is achieved by feeding the network with a delayed version of the past observations, commonly referred to as a delay vector or tapped delay line. These networks do not try to achieve credit assignment back through time, but instead use the previous state as part of the current input. Such a simple approach may be seen as a natural extension to feedforward networks in much the same way that ARIMA models generalize AR models. A network with a rich representation of past outputs is a fully connected RNN and is known as the Williams–Zipser network (NARX network) \cite{r25}, \cite{r26}.

\begin{figure}[t]
\includegraphics[width=.95\columnwidth]{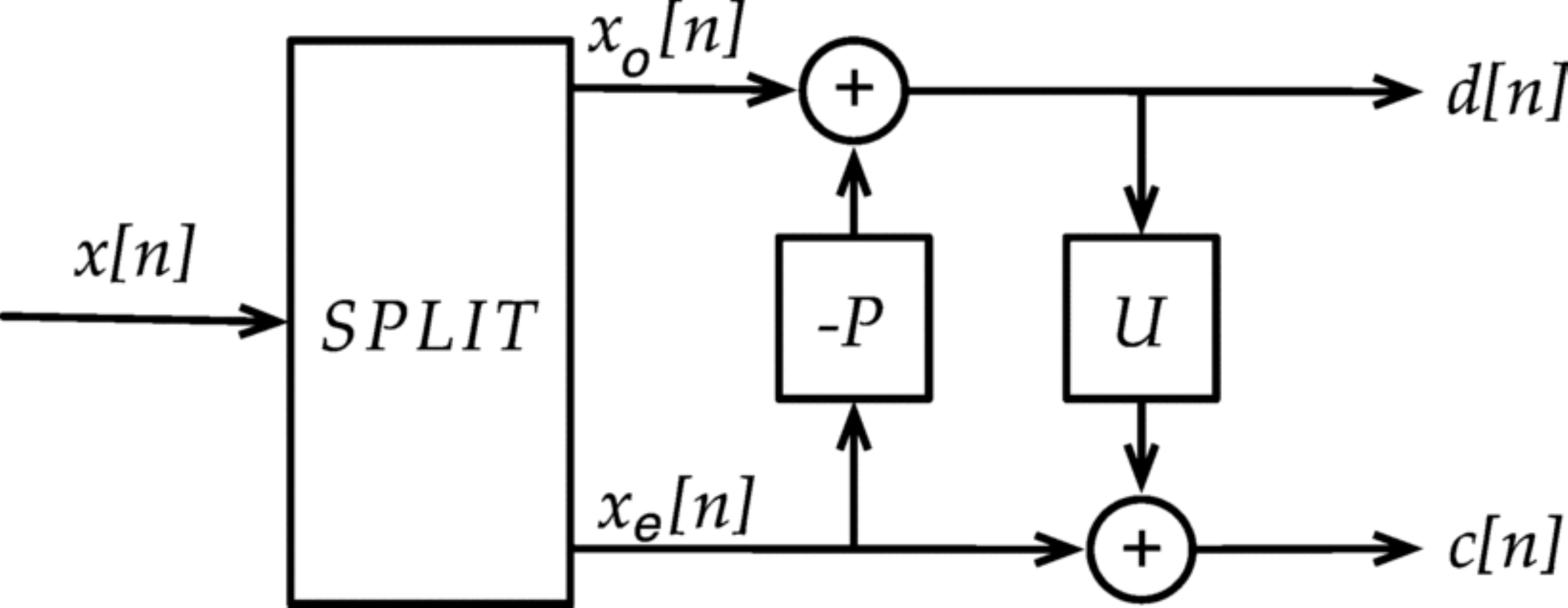}
\caption{Lifting stage: split, predict, and update.}
\label{f1}
\end{figure}

We now give a brief introduction to this architecture. This network consists of two or three layers: the input layer, the processing layer, and the output layer. For each neuron $i$, $i=1,2,\ldots,N$, the elements $n_j$, $j=1,s,\ldots,p+N+1$  of the input vector $\mathbf{u}$ to a neuron as in (\ref{e6}) are weighted and then summed to produce an internal activation function of a neuron $\mathbf{v}$ as in (\ref{e7}), which is finally fed through a nonlinear activation function $\phi$  as in (\ref{e8}) to form the output of the $i^{th}$ neuron $y_i$, as in (\ref{e9}). The function $\phi$ is a monotonically increasing sigmoid function with slope $\beta$, such as, for instance, the logistic function. At the time instant $k$, for the $i^{th}$ neuron, its weights form a $(p+N+1)\times 1$ dimensional weight vector $w(k)=[w_{i,1}(k),\ldots,w(k)=[w_{i,p+N+1}(k)]$ , where $p$ is the number of external inputs and $N$ is the number of feedback connections. One additional element of the weight vector $\mathbf{W}$ is the bias input weight. The feedback consists of the delayed output signals of the RNN. The following equations fully describe the RNN \cite{r27}:
\begin{equation}
\begin{array}{r}
{\bf u}^{T}_{i}(k)=[s(k-1),\ldots,s(k-p),1,\ldots,y_{1}(k-1),~\\
y_{2}(k-1),\ldots,y_{N}(k-1)]
\end{array}
\label{e6}
\end{equation}
\begin{equation}
v_{i}(k)=\sum\limits_{l=1}^{p+N+1}w_{i,l}(k)u_{l}(k)
\label{e7}
\end{equation}
\begin{equation}
\phi (v)={{1}\over{1+e^{^{-\beta v}}}}
\label{e8}
\end{equation}
\begin{equation}
y_{i}(k)=\phi(v_{i}(k)),\qquad i=1,2,\ldots,N
\label{e9}
\end{equation}
Real-time recurrent learning (RTRL) based training of the RNN is achieved upon minimizing the instantaneous squared error at the output of the first neuron of the RNN \cite{r28}, which can be expressed as
\begin{equation}
\min\left({\frac{1}{2}}e^{2}(k)\right)=\min\left({\frac{1}{2}}\left[s(k)-y_{1}(k)\right]^{2}\right)
\label{e10}
\end{equation}
where $e(k)$  denotes the error at the output $y_1$ of the RNN and $s(k)$  is the teaching signal. Hence, the correction for the $l^{th}$ weight of neuron $n$ at the time instant $k$ can be derived as follows:
\begin{equation}
\Delta w_{n,l}(k)=-{\frac{\eta}{2}}{\frac{\partial}{\partial w_{n,l}(k)}}e^{2}(k)=-\eta e(k){\frac{\partial e(k)}{\partial w_{n,l}(k)}}
\label{e11}
\end{equation}
Since the external signal vector  does not depend on the elements of , the error gradient becomes \cite{r28}, \cite{r25}

\begin{equation}
{\frac{\partial e(k)}{\partial w_{n,l}(k)}}e^{2}(k)={\frac{\partial y_{1}}{\partial w_{n,l}(k)}}
\label{e12}
\end{equation}
A triple-indexed set of variables $\{\pi_{n,l}^j(k)\}$ can be introduced to characterize the RTRL algorithm for the RNN as
\begin{equation}
\pi^{j}_{n,l}={{\partial y_{j}(k)}\over{\partial w_{n,l}}},\qquad 1\leq j,\;\; n\leq N,\;\; 1\leq l\leq p+1+N
\label{e13}
\end{equation}
which is used to compute recursively the values of $\pi_{n,l}^j(k)$ for every time step $k$  and all appropriate $j$, $n$ and $l$. A detailed derivation of the RTRL algorithm can be found in \cite{r25}, \cite{r26}, and \cite{r28}.

\begin{table}
\centering
\small
\caption{Number of neurons ($2N$) used for the different kinds of wavelets, the relative RMS, the correlation coefficient ($\Gamma$) between the experimental and the predicted data, and the convergence epochs}
\label{t1}
\begin{tabular}{|l|| c | c | c | c |}
\hline
Kind & $2N$ & RMS & $\Gamma$ & Epochs\\ \hline
Biort. 2.8 & 6 & $< 3$ \% & 0.9987 & 5000 \\  \hline
Biort. 3.7 & 10 & $< 1$ \% & 0.9991 & 8000 \\  \hline
Biort. 3.9 & 10 & $< 1$ \% & 0.9990 & 10000 \\  \hline
Coiflet 2 & 12 & $< 1$ \% & 0.9988 & 11000 \\  \hline
Daub. 4 & 8 & $< 3$ \% & 0.9985 & 6000 \\  \hline
Daub. 6 & 12 & $< 1$ \% & 0.9985 & 6000 \\  \hline
Daub. 8 & 12 & $< 1$ \% & 0.9988 & 12000 \\  \hline
Simlet 4 & 12 & $< 1$ \% & 0.9986 & 10000 \\  \hline
Simlet 7 & 12 & $< 1$ \% & 0.9988 & 9000 \\  \hline
\end{tabular}
\end{table}

\section{Proposed structure of prediction based on Second-Generation Wavelet and RNN}

Now we present the lifting scheme as a simple general construction of second-generation wavelets. The basic idea, which inspired its name, is to start with a very simple or trivial multiresolution analysis and gradually work one's way up to a multiresolution analysis with particular properties. The lifting schema allows one to custom-design the filters needed in the transform algorithms to the situation at hand. In this sense, it provides an answer to the algebraic stage of a wavelet construction \cite{r29}. Wavelet functions $\psi_{j,m}$ are traditionally defined as the dyadic translations and dilations on one particular $L^2(\mathbb{R})$, i.e., the mother wavelet $\psi:\psi_{j,m}=\psi(2^j x - m)$. We refer to such wavelets as first-generation wavelets. In this paper, we introduce a more general setting where the wavelets are not necessarily translations and dilations of each other but still enjoy all the powerful properties of first-generation wavelets. These wavelets are referred to as second-generation wavelets.

\begin{figure}[t]
\includegraphics[width=.95\columnwidth]{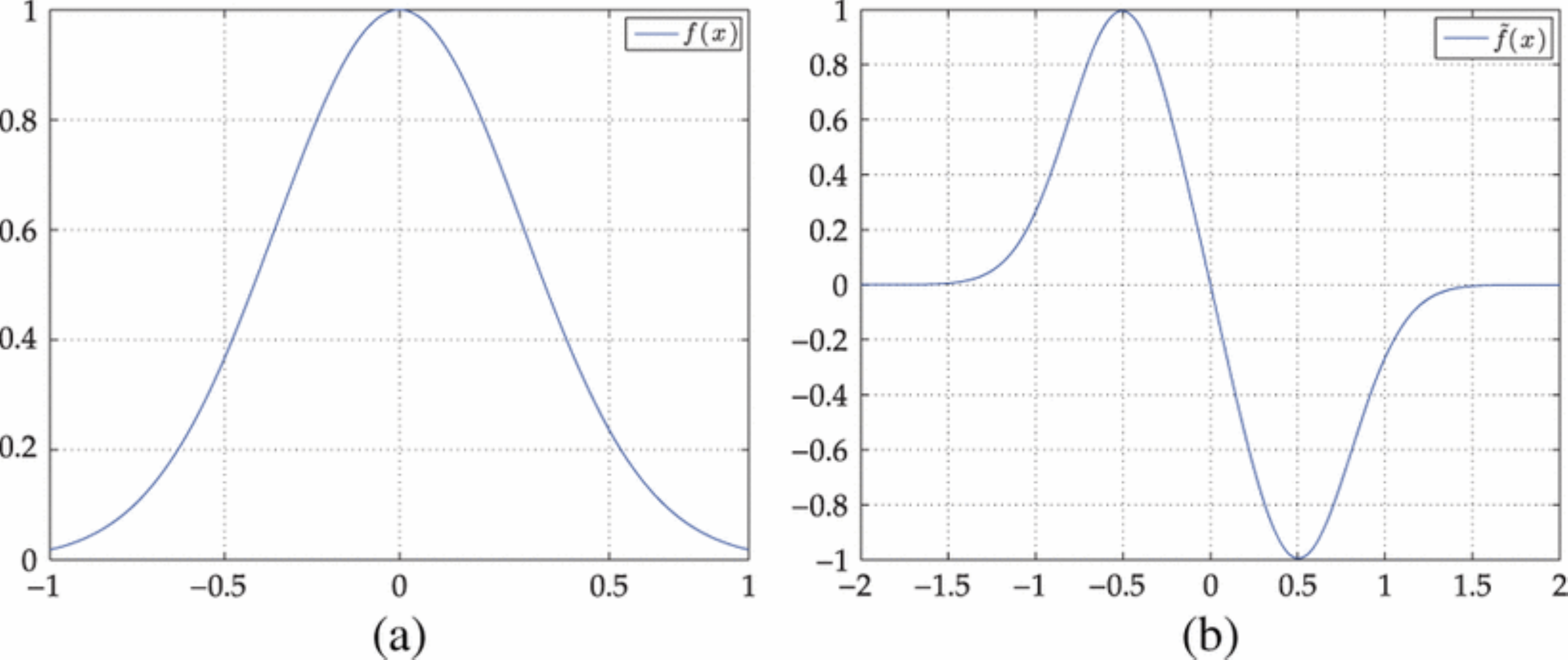}
\caption{ (a) RBF transfer function $f(x)$ and (b) relative wavelet function $\tilde{f}(x)$.}
\label{f2}
\end{figure}

\subsection{Basics of Second-Generation Wavelet Theory}

We present the lifting scheme, which is a simple but quite powerful tool, to construct second-generation wavelets. Lifting is made by a space-domain construction of biorthogonal wavelets developed in \cite{r21}. It consists of the iteration of the following three basic operations: The split consists in dividing the original data into two disjointed subsets. For example, we will split the original dataset $x[n]$ into $x_e [n]=x[2n]$, which are the even indexed points, and $x_o [n]=x[2n+1]$, which are the odd indexed points. The prediction consists in generating the wavelet coefficients $d[n]$ as the error in predicting $x_o [n]$ from $x_e [n]$ using the prediction operator $P$

\begin{equation}
d[n]=x_o [n] - Px_e [n]
\label{e14}
\end{equation}

The update consists in combining $x_e [n]$ and $d[n]$ to obtain scaling coefficients $c[n]$ which represent a coarse approximation to the original signal $x[n]$. This is accomplished by applying an updating operator $U$ to the wavelet coefficients and adding to $x_e [n]$

\begin{equation}
c[n]=x_e [n] + U d[n]
\label{e15}
\end{equation}

These three steps form a lifting stage. Iteration of the lifting stage on the output $c[n]$ creates the complete set of discrete wavelet transform (DWT) scaling and the wavelet coefficients $c^j[n]$ and $d^j[n]$. The lifting steps are easily inverted even if $P$ and $U$ are nonlinear or space-varying. Rearranging (\ref{e14}) and (\ref{e15}), we have 
\begin{equation}
\left \{
\begin{array}{l}
x_e[n]=c[n]-Ud[n]\\
x_o[n]=d[n]+Px_e[n]
\end{array}
\right .
\label{e16}
\end{equation}

The described lifting scheme also leads to a fast in-place calculation of the wavelet transform, i.e., an implementation that does not require auxiliary memory. In the lifting framework of Fig.~\ref{f1}, the update structure depends on the predictor structure. Hence, if $P$ is space-varying or nonlinear, then so is $U$, and the design procedure becomes unwieldy. A crafty detour around this problem is to perform the update step first, followed by the prediction. The relevant equations then become 

\begin{equation}
\left \{
\begin{array}{l}
c[n]=x_e [n]+Ux_o[n]\\
d[n]=c[n]-Px_e[n]
\end{array}
\right .
\label{e17}
\end{equation}

After designing an update filter to preserve the first $\tilde{N}$ low-order polynomials in the data, we can apply any space-varying or nonlinear predictor without affecting the coarse approximation $c[n]$. Since the update and predict lifting stage creates $c[n]$ prior to $d[n]$, the prediction operator can be designed to optimize the performance criteria in addition to polynomial suppression capability \cite{r29}.

\begin{figure}[t]
\includegraphics[width=.95\columnwidth]{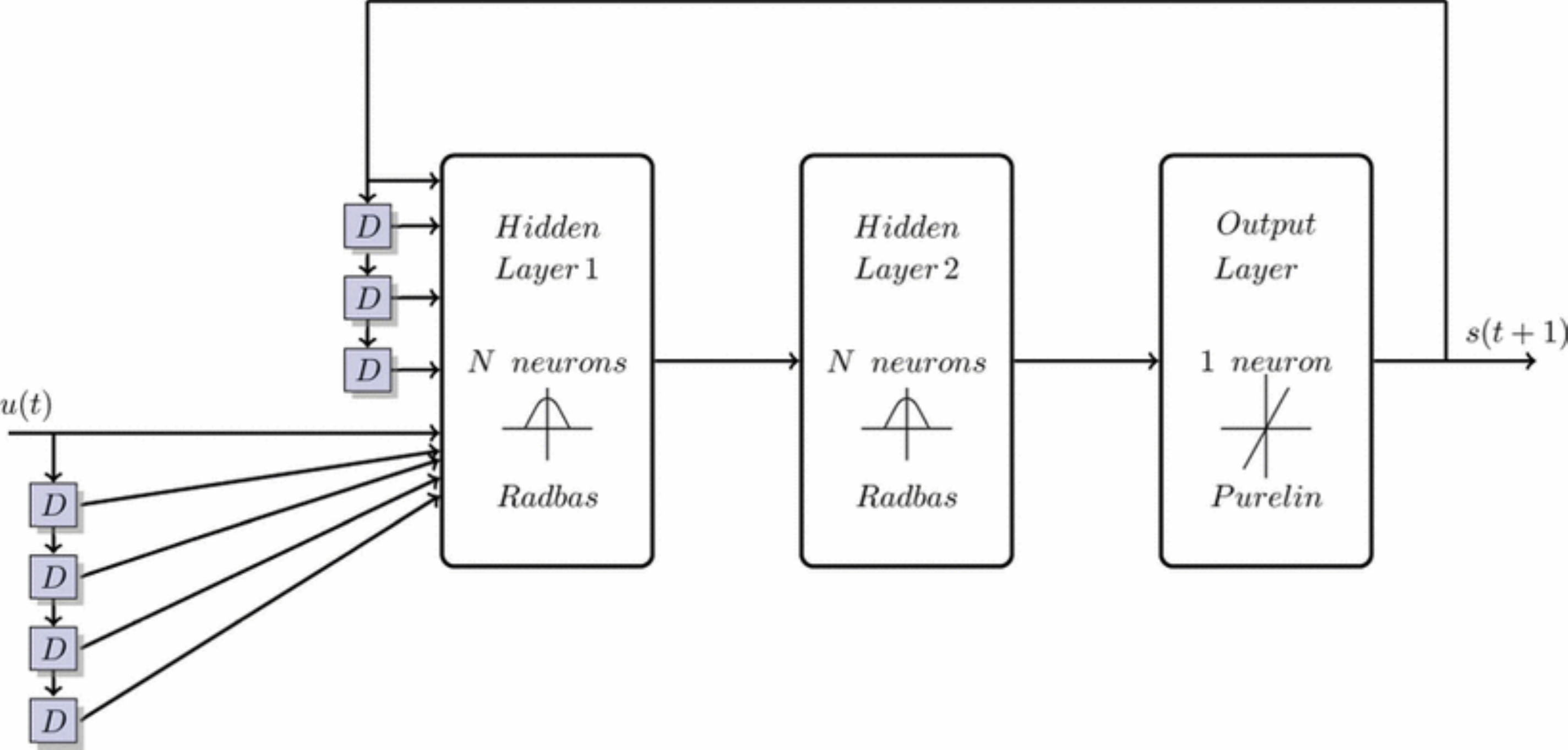}
\caption{Selected RNN topology for the buildup of $P$ and $U$.}
\label{f3}
\end{figure}

\subsection{Proposed Novel Buildup of P and U Based on RNN}

RNN has already been proven to be able to exploit the intrinsic features of time series in order to predict its temporal evolution \cite{r30}, \cite{r31}. In the aim to design a forecast neural network, wavelet coefficients as input set give a better and efficient expression of these intrinsic features, packing in a few significant coefficients all the energy (and information) carried by the input signals. In \cite{r32} and \cite{r33}, the authors show a properly designed hybrid neuro-wavelet recurrent network that is also able to execute wavelet reconstruction and prediction of a signal. In this paper, we derive a lifting and updating construction based on a polynomial signal suppression and preservation argument. Exploiting the generalization and prediction properties of the RNNs, we realize both the operators $P$ and $U$, thereby tailoring the relative structures to do the lifting and predicting stages. More generally, we let the signal itself dictate the structure of the predictor. In a scale-adapted transform, we adapt the predictor in each lifting stage in order to match the signal structure at the corresponding scale. The basic idea is to use an RNN to adapt the predictor to the signal. This optimization produces predictors that can match both polynomial and nonpolynomial signal structures. The optimization itself is a straightforward $N$-dimensional constrained least squares problem. The constraint is that we require the predictor to suppress $N^{th} \leq M^{th}$ order polynomials, they being related to the well-known vanish moments characterizing the various wavelet systems and summarized in Table~\ref{t1}. Now let $x_o$ denote the odd-indexed data we wish to predict, and let $X_e:[X_e]_{n,k}=x_e[n-k]$ be a matrix composed of the even-indexed data used in the prediction. The vector of prediction errors then is given by 

\begin{equation}
e = x_o - X_e p
\label{e18}
\end{equation}

Our objective is to find the prediction coefficients that minimize the sum of squared prediction errors $e^T e$ while satisfying the $N \leq M$ polynomial constraints. This we solve as 

\begin{equation}
\min\Vert x_{o}-X_{e}p\Vert^{2}
\label{e19}
\end{equation}

The optimal prediction coefficients for this constrained least squares problem can be found efficiently. We call the proposed system WRNN. In fact, it is able to reconstruct a signal from wavelet coefficients, and also to predict these wavelet coefficients aiming to reconstruct and forecast the signal. To obtain this behavior, some rules have to be applied during the design and implementation stages. For reasons that will become clear later, all the hidden layers have a pair neuron number; also, to permit in sequence the wavelet coefficient exploitation and the signal reconstruction, a double hidden layer is required in the proposed architecture. As for the hidden layers, the neuron's activation function (transfer function) has to simulate a wavelet function. It is not possible to implement a wavelet function itself as transfer function for a forecast-oriented time-predictive neural network, because wavelets do not verify some basic properties such as the absence of local minima and do not provide by themselves a sufficiently graded response \cite{r34}. In the existing range of possible transfer functions, only some particular classes approximate the functional form of a wavelet.

In this paper, RBFs are chosen as transfer functions for the selected RNNs, as shown in Fig.~\ref{f2}. Indeed, these particular kinds of functions well describe half of a wavelet in first approximation, even though these do not verify the properties shown by (\ref{e2}) and (\ref{e3}). Anyway, after scaling, shifting, and repetition of the chosen activation function, it is possible to obtain several mother wavelet filters. Let $f:[-1;1]\rightarrow\mathbb{R}^{+}$ be the chosen transfer function; then it verifies all the properties of a wavelet function 

\begin{equation}
\tilde{f}(x+2k)= \left \{
\begin{array}{lr}
+f(2x+1) & ~x \in [-1,0] \\
-f(2x-1) & ~x \in [0,+1]
\end{array}
\right . ~~~ \forall~k \in \mathbb{Z}
\label{e20}
\end{equation}

So it is possible for the selected neural networks to simulate a wavelet by using the radbas function defined in the $[-1;1]$ real domain. It is indeed possible to verify that 

\begin{equation}
\int_{2h+1}^{2k+1} \tilde{f}(x) dx = 0 \quad \forall~ h < k \in \mathbb{Z}
\label{e21}
\end{equation}

It was shown that, in order to simulate a wavelet function, these chosen transfer functions have to be symmetrically periodic to emulate a wavelet. This is the reason why we choose a pair of neurons with the aim of having the same number of positive and negative layer weights in the reconstruction layer. Theoretically, if this happens, then the neuron pairs of the second layer emulate exactly a reconstruction filter. Although this is a theoretical schema, there are strong reasons for the weights in this experimental setup to have a nonzero sum because the neural network beyond to perform the inverse wavelet transform must perform also the signals prediction.

Initially, several wavelet decompositions were used to obtain an input vector of wavelet coefficients, with the aim to study the capability of the selected neural network to reconstruct and predict the signal simulating different kinds of wavelet functions. Different kinds of topology and size variations were also implemented to select the better performing neural network design. The implemented wavelet decompositions permit the location of the coefficient bands related to the timescale relative to the prediction goals. By thresholding to zeros the bands unrelated to the selected timescales, the resulting coefficients and residuals carry relevant information for the predictions. These wavelet coefficients were then provided as input ($\vec{u}_i(t)$) to the system. The selected neural network is composed of two hidden layers of 16 neurons and a single output neuron. The wavelet decomposition of the time series is given as $N \times 4$ input vectors at time $t_0$ with a three-step delay and a one-step output feedback to predict the output signals $s(t_0 + 1)$. To improve the generalization of the selected RNN, we have used the common method called early stopping. The considered RNN topology for the proposed novel buildup of $P$ and $U$ is shown in Fig.~\ref{f3}.

\begin{figure*}[t]
\centering
\begin{minipage}{.370\textwidth}
\includegraphics[width=\textwidth]{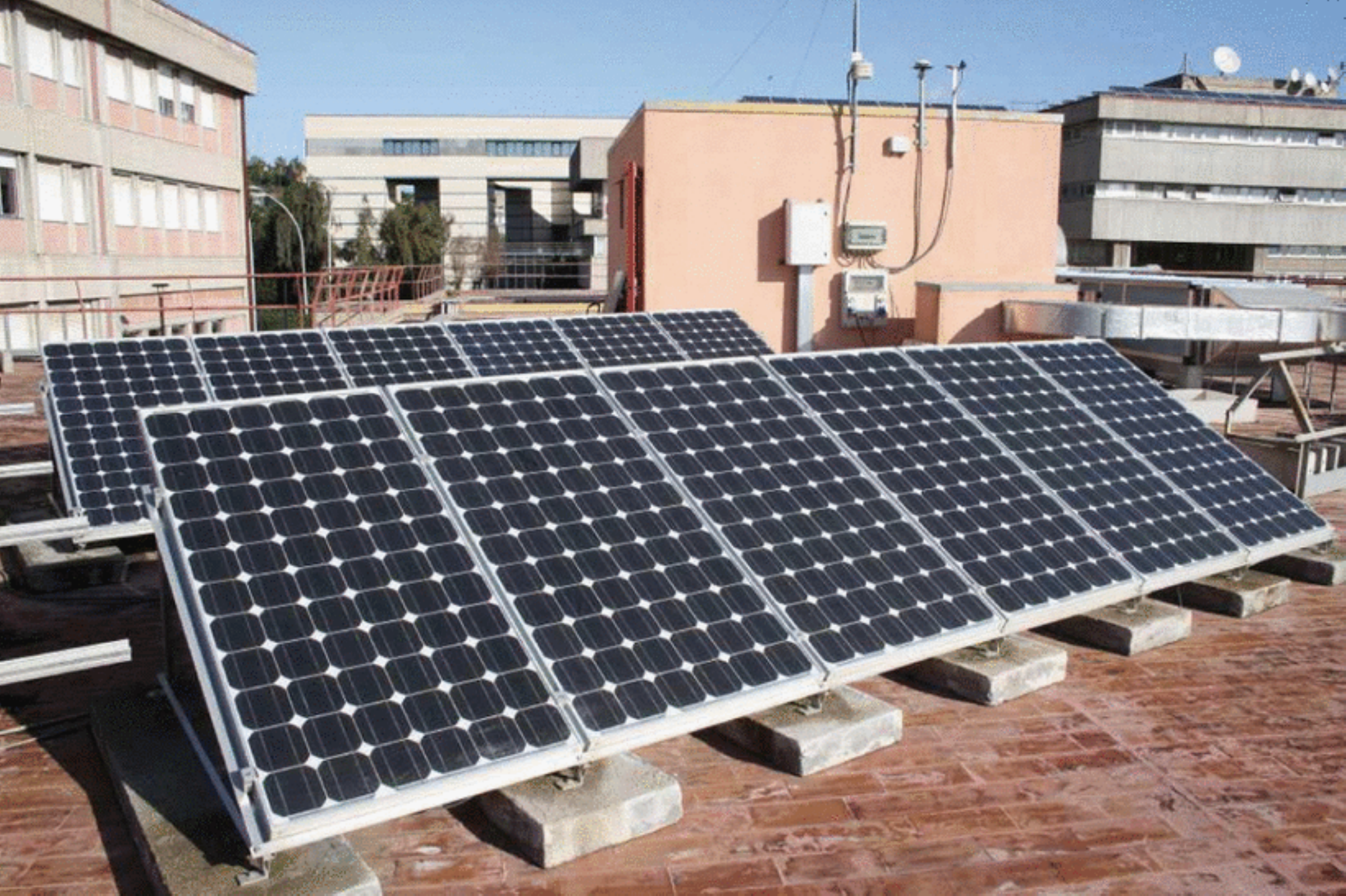}
\caption{PV modules.}
\label{f4}
\end{minipage}
~~~~~
\begin{minipage}{.530\textwidth}
\includegraphics[width=\textwidth]{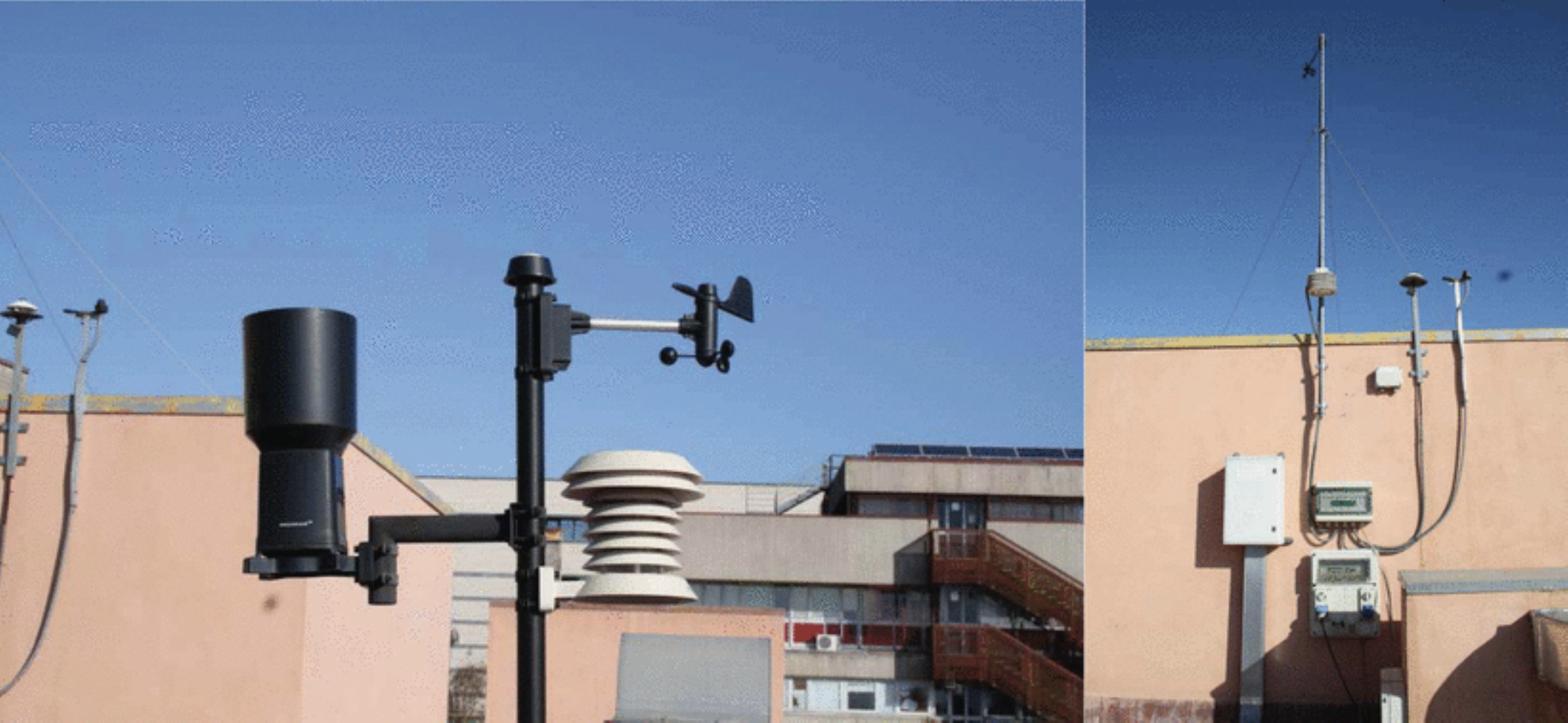}
\caption{Metereological data instrumentation.}
\label{f5}
\end{minipage}
\end{figure*}

\section{Experimental Setup}
The IDRILAB laboratories of the University of Catania provided the experimental study of the electrical behavior of PV modules and strings. The facility incorporates meteorological instrumentation with a data acquisition system, and electronic loads to obtain the plot of the current versus voltage (I–V curves) of the PV modules shown in Fig.~\ref{f4}. Precision spectral pyranometers (thermopile and photodiode) are used to measure the total solar radiation. A three-cup anemometer and wind vane assembly is used to measure the wind speed and wind direction, as shown in Fig.~\ref{f5}. Ambient temperature is measured using a perforated-tip type-T thermocouple sensor enclosed in a naturally ventilated multiplate radiation shield. The meteorological data are provided in wireless mode by means of a ZigBee interface in conjunction with a software platform based on the functional schema of the meteorological data station. The meteorological variables are stored in a MySQL database. The power supply of the sensor module has two voltage levels: 5 VDC for the anemometer, ambient temperature, relative humidity, and photodiode pyranometer; and $\pm 12$ VDC  for the termopile pyranometer. The processed data was gathered by a WSENS metero unit from the following sensors.

\begin{enumerate}
\item{Wind speed}
	\begin{itemize}
	\item Type: Polycarbonate cups with magnetic switch.
    	\item Range: 3--241 km/h.
    	\item Accuracy: 3 km/h.
    	\item Resolution: 1 km/h.
	\end{itemize}
\item{Temperature}
	\begin{itemize}
	\item Type: Bandgap with digital output.
    	\item Range: $-40^{\circ}$C to $123.8^{\circ}$C.
	\item Resolution: $0.01^{\circ}$.
    	\item Accuracy: $1^{\circ}$ in the range $−20^{\circ}$C to $50^{\circ}$C.
	\end{itemize}
\item{Relative humidity}
	\begin{itemize}
	\item Type: Capacitive polymer with digital output.
    	\item Range: 0\%--100\%.
    	\item Resolution: 0.03\%.
    	\item Accuracy: 2\% from 10\%--90\%.
	\end{itemize}
\item{Solar irradiation}
	\begin{itemize}
	\item Type: Thermopile pyranometer.
    	\item Range: 0--1400 W/$\mbox{m}^2$
    	\item Resolution: 1\%.
    	\item Accuracy: 2  W/$\mbox{m}^2$
	\end{itemize}
\end{enumerate}

A long-term data survey was made from June 2007 to November 2008 at the IDRILAB Laboratory, University of Catania. During the acquisition period, the data were registered as described in the following. Wind velocities were recorded using a polycarbonate cup with magnetic switch, capable of measurements in the range 3–241 km/h with 5\% accuracy. RH measurements were made with a capacitive polymer with digital output, providing 2\% accuracy in the range 10\%-90\%. For the temperature measurements, we used a digital bandgap in the range $-20^{\circ}$C to $50^{\circ}$C with $1^{\circ}$ uncertainty. The solar radiation on the horizontal plane was measured by a pyrometer.

\section{Simulation results}
\begin{figure*}[thb]
\includegraphics[width=.95\textwidth]{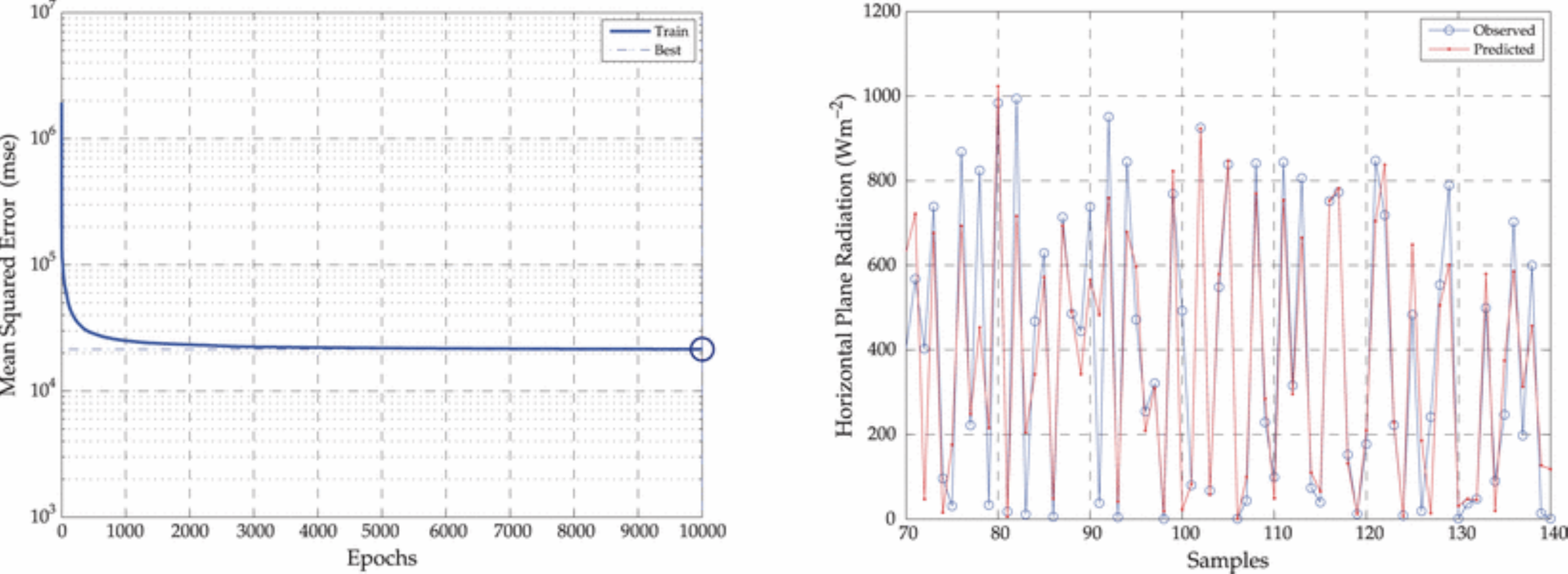}
\caption{Wavelet biortogonal 2.8 decomposition set.}
\label{f6}
\end{figure*}

\begin{figure*}[thb]
\includegraphics[width=.95\textwidth]{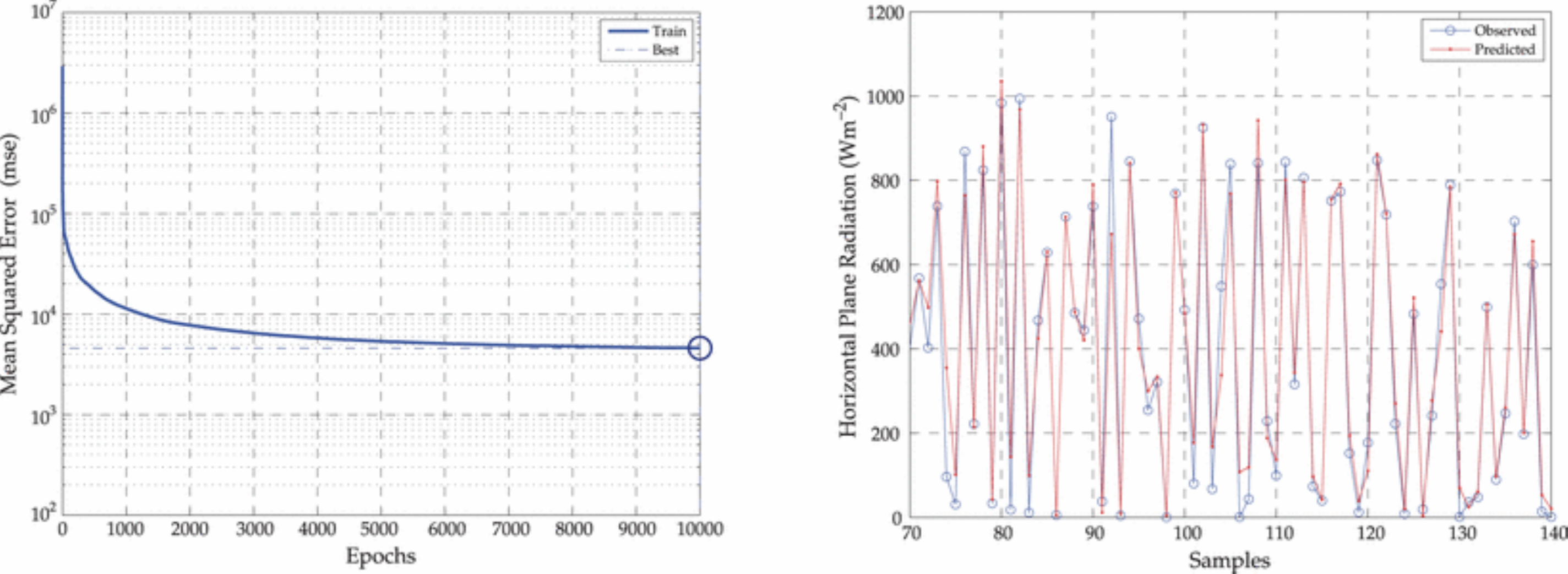}
\caption{Wavelet biortogonal 3.7 decomposition set.}
\label{f7}
\end{figure*}

\begin{figure*}[thb]
\includegraphics[width=.95\textwidth]{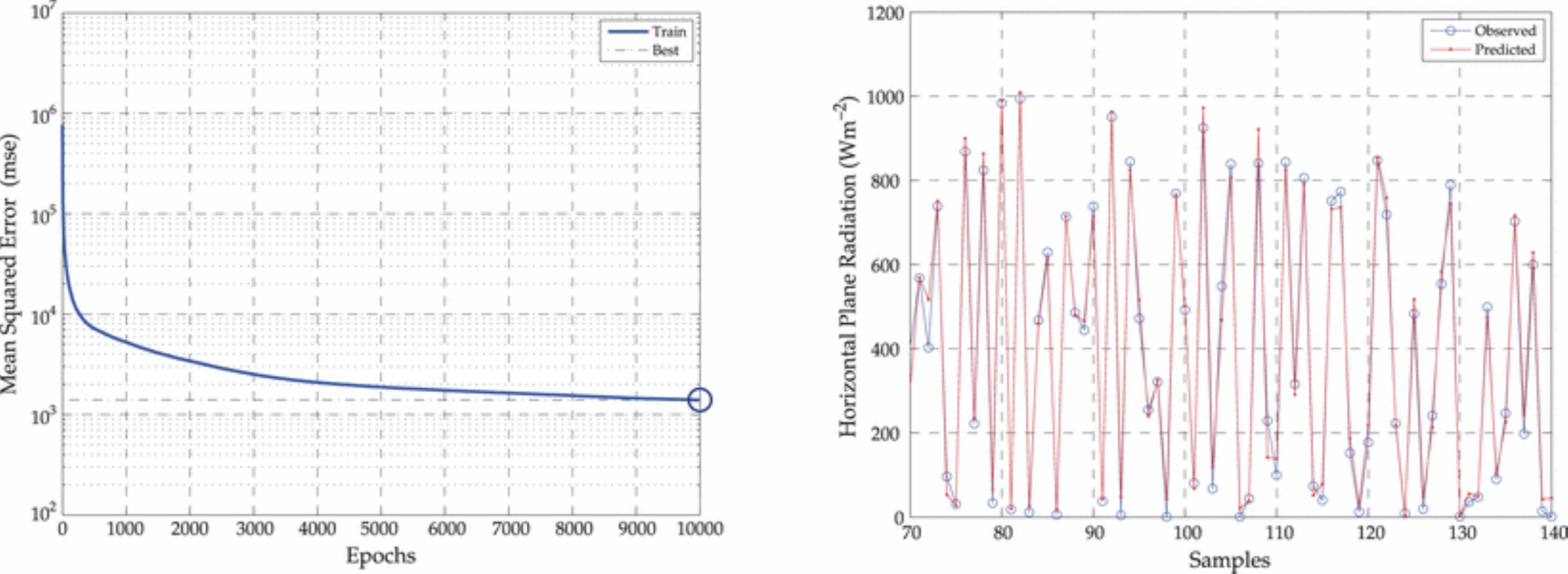}
\caption{Wavelet biortogonal 3.9 decomposition set.}
\label{f8}
\end{figure*}

\begin{figure*}[thb]
\includegraphics[width=.95\textwidth]{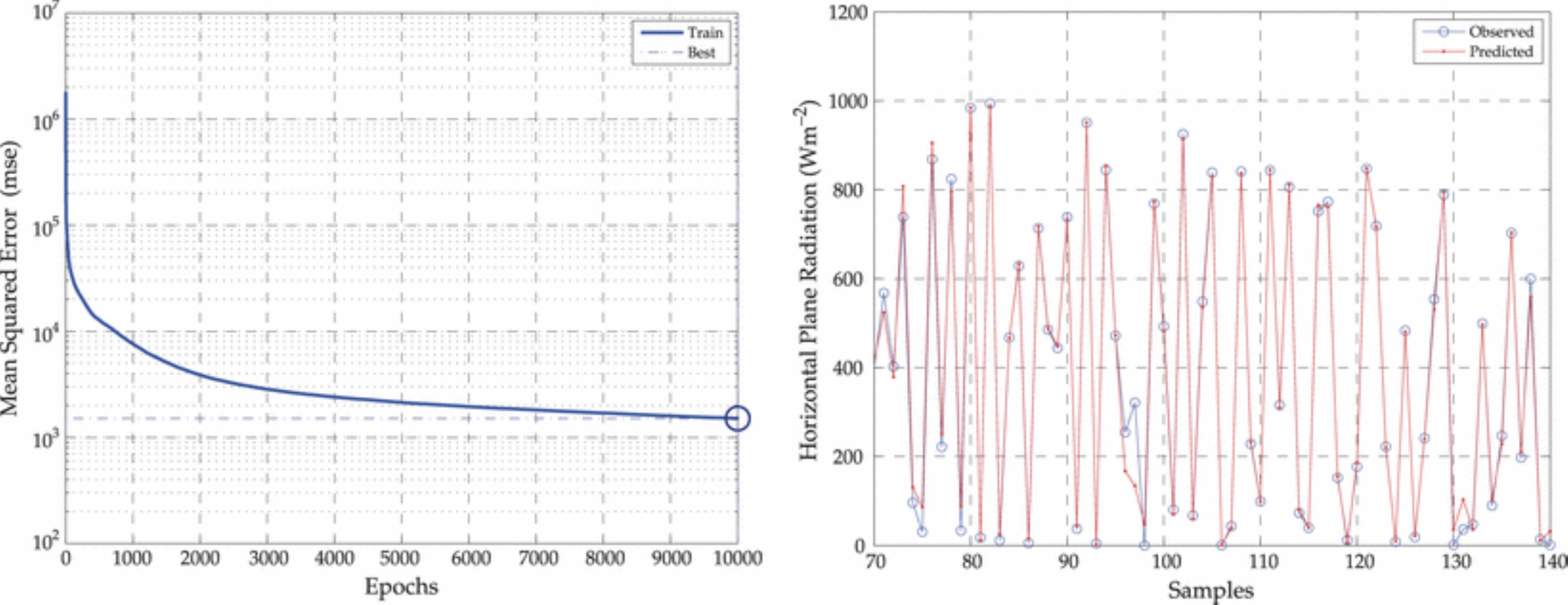}
\caption{Wavelet coiflet 2 decomposition set.}
\label{f9}
\end{figure*}

\begin{figure*}[thb]
\includegraphics[width=.95\textwidth]{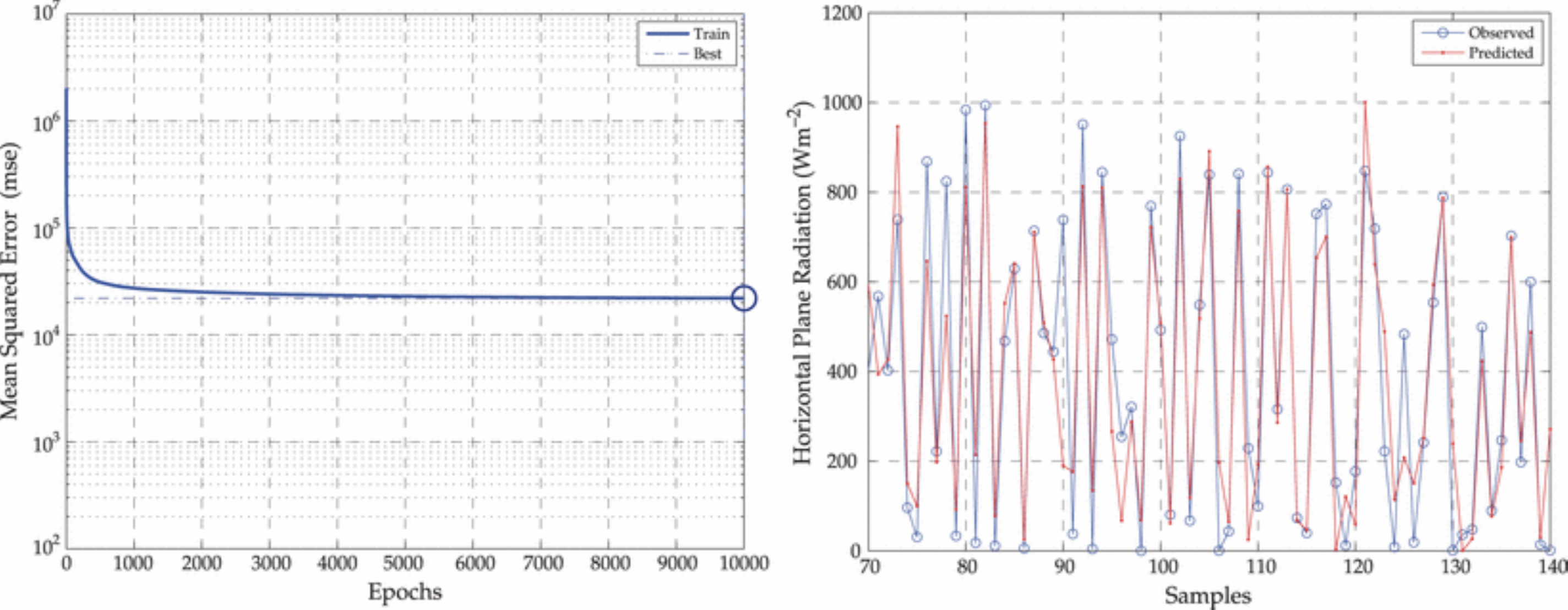}
\caption{Wavelet Daubechies 4 decomposition set.}
\label{f10}
\end{figure*}

\begin{figure*}[thb]
\includegraphics[width=.95\textwidth]{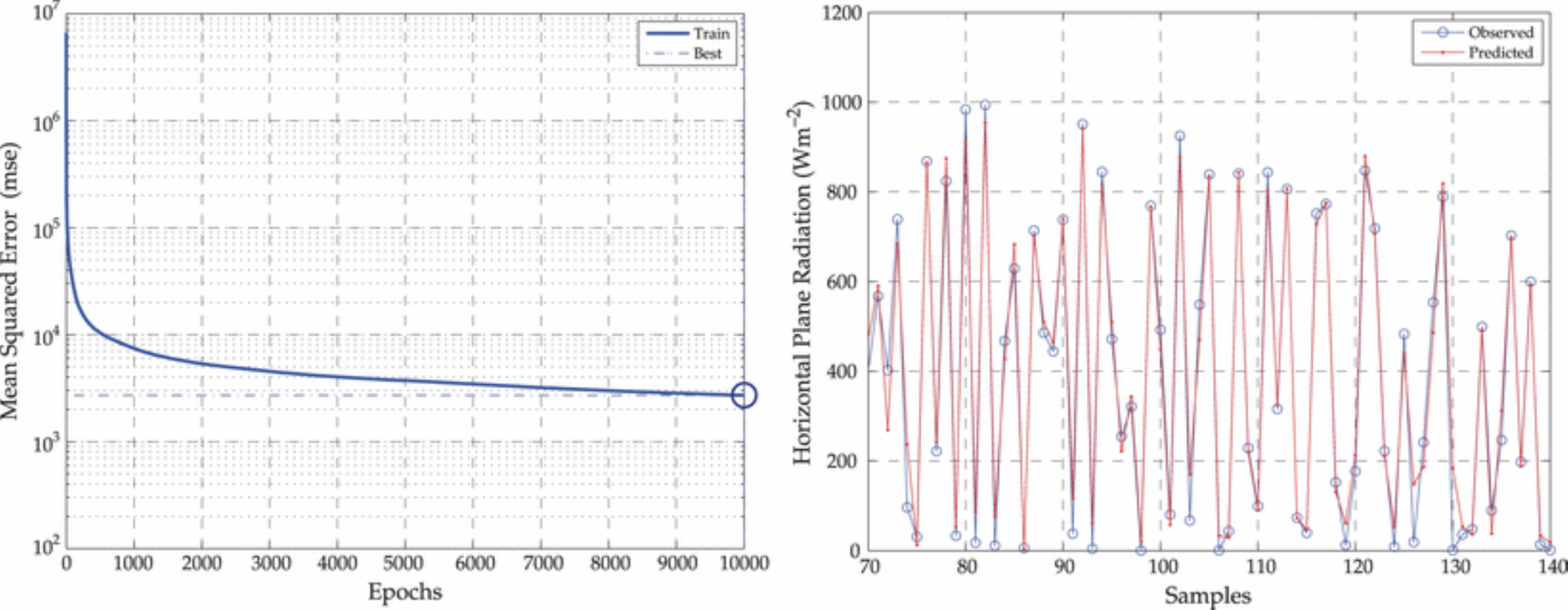}
\caption{Wavelet Daubechies 6 decomposition set.}
\label{f11}
\end{figure*}

\begin{figure*}[thb]
\includegraphics[width=.95\textwidth]{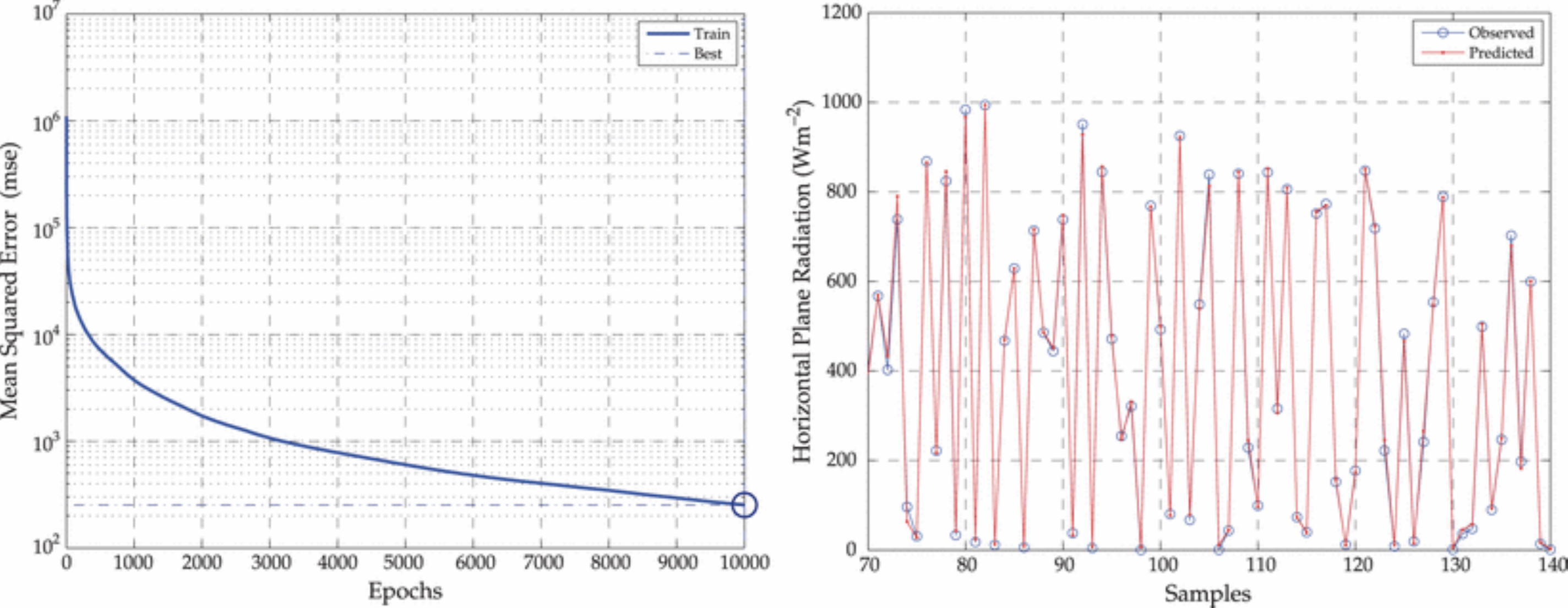}
\caption{Wavelet Daubechies 8 decomposition set.}
\label{f12}
\end{figure*}

\begin{figure*}[thb]
\includegraphics[width=.95\textwidth]{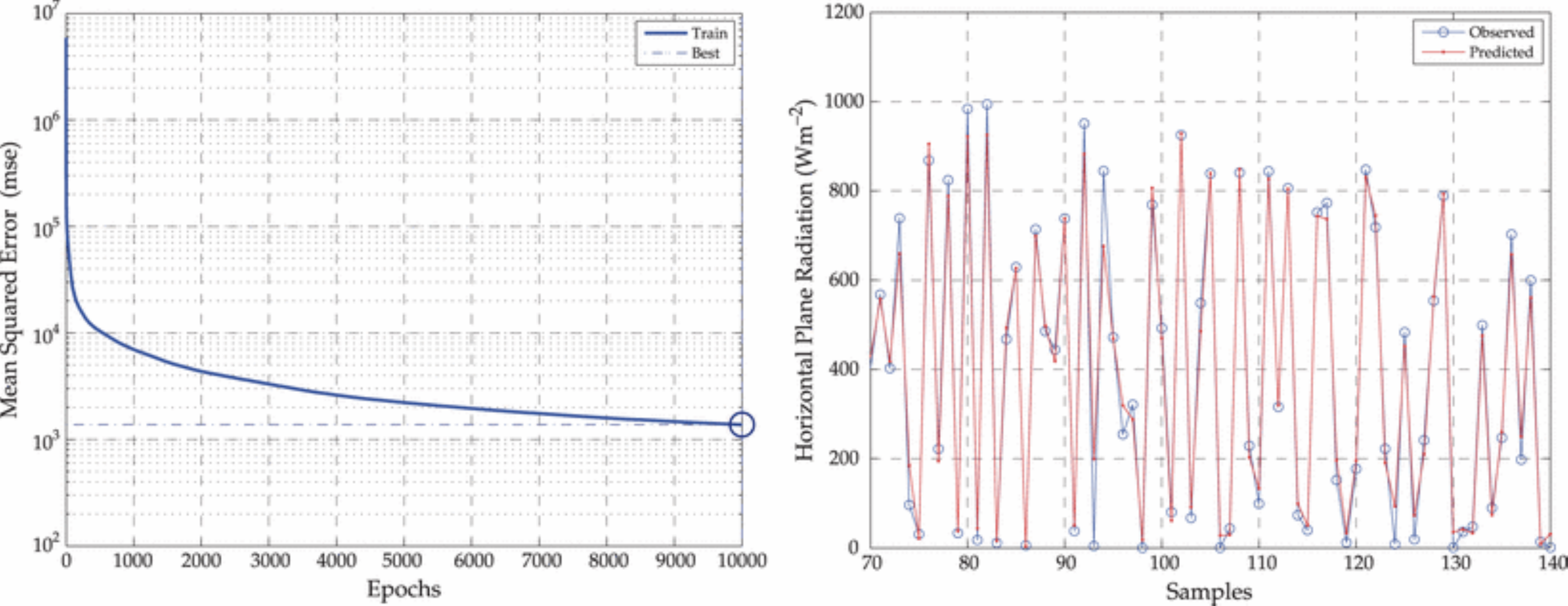}
\caption{Wavelet Simlet 4 decomposition set.}
\label{f13}
\end{figure*}

\begin{figure*}[thbn]
\includegraphics[width=.95\textwidth]{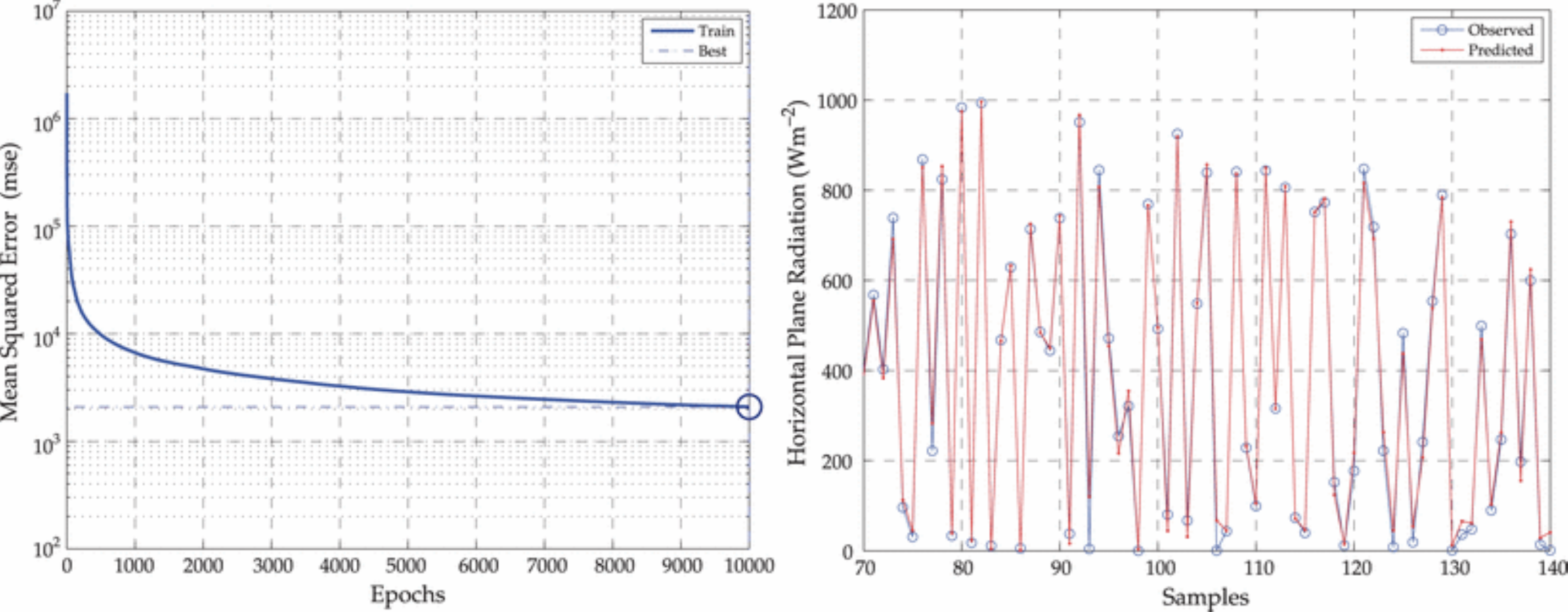}
\caption{Wavelet Simlet 7 decomposition set.}
\label{f14}
\end{figure*}

The experimental data were collected at sampling intervals of 10 min and automatically stored as synchronized time series by the information infrastructure managed by the laboratory staff. The collected input dataset was at first decomposed using the wavelet decompositions reported in Table~\ref{t1}. The wavelet scale was chosen to have a dyadic expansion so that the first band could be representative of 2-day time steps. This was made for the temperature, RH, and wind speed time series. The measured horizontal plane solar radiation ranged from 0 to 1300 $\mbox{Wm}^2$. The input pattern was composed of 12 element vectors (4 elements of 3 types of measurements). The input elements were the $a_1(t_0)$ residues, and the $d_1(t_0)$, $d_2(t_{0^{-}})$, and $d_2(t_{0^{+}})$ details. The neural networks were trained to predict 1-D output signals $s(t_1)$ at about two days in the future. In Table~\ref{t1}, we report the main features of the different networks implemented for the different kinds of wavelets relative to the simulation results depicted in Figs.~\ref{f6}--\ref{f14}. The data series are very long, therefore, in order to plot in a clear manner the obtained solar radiation prediction, only a small part is presented Figs.~\ref{f6}--\ref{f14}, and the data sample was chosen in an arbitrary manner. The quality of prediction was the same for the time data series as a whole. Here, the time refers to 10000 epochs. As shown in Table~\ref{t1}, the number of neurons doubles the number of vanishing moments plus a variable number. In fact, the synthesis filter affects this variable because of the use of this filter by the network for the prediction in the wavelet domain.


\section{Conclusions}
Efficient development and utilization of solar energy and accurate knowledge of the availability and variability of solar radiation intensity, both in time and spatial domains, are very critical issues in renewable energy systems, such as PV modules. The experimental setup available at the University of Catania provided the latter information. ANNs have been used for several years for various time-series forecasting applications, but the lack of a systematic neural network modeling procedure and strategy has led to exploring other innovative approaches, especially using wavelet theory and wavelet networks. Several papers have appeared in the recent literature on forecasting problems faced by using neural networks, some of which are listed in \cite{r35}, \cite{r36}, \cite{r37}, \cite{r38}, \cite{r39}. The good performance of our model forecasting has led to the following main conclusions: The novelty of this approach is that the proposed WRNN does not provide the wavelet coefficients coming from the intrinsic information from other input coefficients; indeed, it is able to reconstruct them directly from the sampled signal from band-selected coefficients. Elizondo et al.proposed the use of a feedforward neural network to estimate the measured daily total solar radiation, the predicted daily total solar radiation, the tilt angle of panel efficiency of the PV array, as well as other parameters such as the temperature, precipitation, clear-sky radiation, day length, and day of the year \cite{r12}. Al-Alawi and Al-Hinai \cite{r9} used an ANN to analyze the relationship between the total radiation and climatology variables; this model predicted the total radiation values to a good accuracy of approximately 93\%. The key point of the reported solar radiation forecasting is the novel structure of the prediction which is based on second-generation wavelets and RNN. Guessoum et al. \cite{r20} have used RBF networks to predict solar radiation data. In their work, the input and the output are the solar radiation data corresponding to a particular day and those of the next day. Kalogirou et al. \cite{r11} used an RNN to predict the maximum solar radiation from relative humidity and temperature. The results indicate that the correlation coefficient obtained varied between 98.58\% and 98.75\%. Some authors have proposed particle swarm optimization and adaptive neuro fuzzy inference systems approaches for short-term renewable sources forecasting. From the computational point of view, the proposed approach is more efficient because of the fast convergence of the neural network, as shown by the mean square error (MSE) traces plotted in Figs.~\ref{f6}--\ref{f14}. The proposed forecasting method is robust to data errors. It leads to savings in the inverse wavelet transform calculation. Finally, it represents an innovation from the methodological point of view. The simulation results obtained with the proposed forecasting method show a very low RMS error compared to those obtained by other solar radiation prediction methods based on hybrid neural networks already developed.

\appendices
\section*{Acknowledgment}

This paper has been published in the final and reviewed version on  {\bf IEEE Transactions on Neural Networks and Learning Systems, vol. 23, N. 11, pp. 1805-1815, 2012} \cite{rr}.

\bibliographystyle{IEEEtran}
\bibliography{napoliOA12NNLS1}

\begin{thebibliography}{10}
\providecommand{\url}[1]{#1}
\csname url@samestyle\endcsname
\providecommand{\newblock}{\relax}
\providecommand{\bibinfo}[2]{#2}
\providecommand{\BIBentrySTDinterwordspacing}{\spaceskip=0pt\relax}
\providecommand{\BIBentryALTinterwordstretchfactor}{4}
\providecommand{\BIBentryALTinterwordspacing}{\spaceskip=\fontdimen2\font plus
\BIBentryALTinterwordstretchfactor\fontdimen3\font minus
  \fontdimen4\font\relax}
\providecommand{\BIBforeignlanguage}[2]{{%
\expandafter\ifx\csname l@#1\endcsname\relax
\typeout{** WARNING: IEEEtran.bst: No hyphenation pattern has been}%
\typeout{** loaded for the language `#1'. Using the pattern for}%
\typeout{** the default language instead.}%
\else
\language=\csname l@#1\endcsname
\fi
#2}}
\providecommand{\BIBdecl}{\relax}
\BIBdecl

\bibitem{r1}
M.~Egido and E.~Lorenzo, ``The sizing of stand alone pv-system: a review and a
  proposed new method,'' \emph{Solar Energy Materials and Solar Cells},
  vol.~26, no.~1, pp. 51--69, 1992.

\bibitem{r2}
A.~Mellit, M.~Benghanem, A.~Hadj~Arab, and A.~Guessoum, ``Modelling of sizing
  the photovoltaic system parameters using artificial neural network,'' in
  \emph{Control Applications, 2003. CCA 2003. Proceedings of 2003 IEEE
  Conference on}, vol.~1.\hskip 1em plus 0.5em minus 0.4em\relax IEEE, 2003,
  pp. 353--357.

\bibitem{r3}
S.~Hokoi, M.~Matsumoto, and M.~Kagawa, ``Stochastic models of solar radiation
  and outdoor temperature,'' \emph{ASHRAE Transactions (American Society of
  Heating, Refrigerating and Air-Conditioning Engineers);(United States)},
  vol.~96, no. CONF-9006117--, 1990.

\bibitem{r4}
M.~Nehrir, C.~Wang, K.~Strunz, H.~Aki, R.~Ramakumar, J.~Bing, Z.~Miao, and
  Z.~Salameh, ``A review of hybrid renewable/alternative energy systems for
  electric power generation: Configurations, control, and applications,''
  \emph{Sustainable Energy, IEEE Transactions on}, vol.~2, no.~4, pp. 392--403,
  2011.

\bibitem{r5}
R.~Aguiar and M.~Collares-Pereira, ``Tag: A time-dependent, autoregressive,
  gaussian model for generating synthetic hourly radiation,'' \emph{Solar
  Energy}, vol.~49, no.~3, pp. 167--174, 1992.

\bibitem{r6}
K.~Knight, S.~Klein, and J.~Duffie, ``A methodology for the synthesis of hourly
  weather data,'' \emph{Solar Energy}, vol.~46, no.~2, pp. 109--120, 1991.

\bibitem{r7}
A.~Maafi and A.~Adane, ``A two-state markovian model of global irradiation
  suitable for photovoltaic conversion,'' \emph{Solar \& wind technology},
  vol.~6, no.~3, pp. 247--252, 1989.

\bibitem{r8}
L.~Mora-Lopez and M.~Sidrach-de Cardona, ``Multiplicative arma models to
  generate hourly series of global irradiation,'' \emph{Solar Energy}, vol.~63,
  no.~5, pp. 283--291, 1998.

\bibitem{r9}
S.~Al-Alawi and H.~Al-Hinai, ``An ann-based approach for predicting global
  radiation in locations with no direct measurement instrumentation,''
  \emph{Renewable Energy}, vol.~14, no.~1, pp. 199--204, 1998.

\bibitem{r10}
Y.~Kemmoku, S.~Orita, S.~Nakagawa, and T.~Sakakibara, ``Daily insolation
  forecasting using a multi-stage neural network,'' \emph{Solar Energy},
  vol.~66, no.~3, pp. 193--199, 1999.

\bibitem{r11}
S.~A. Kalogirou, S.~Michaelides, and F.~Tymvios, ``Prediction of maximum solar
  radiation using artificial neural networks,'' in \emph{Proc. World Renew.
  Energy Congr.}, 2002, pp. 353--357.

\bibitem{r12}
D.~Elizondo, G.~Hoogenboom, and R.~W.~M. Clendon, ``Development of a neural
  network model to predict daily solar radiation,'' \emph{Agricultural and
  Forest Meteorologyy}, vol.~71, no. 1-2, pp. 115--132, 1994.

\bibitem{r13}
I.~Daubechies, ``The wavelet transform, time-frequency localization and signal
  analysis,'' \emph{Information Theory, IEEE Transactions on}, vol.~36, no.~5,
  pp. 961--1005, 1990.

\bibitem{r14}
S.~A. Imhoff, D.~Y. Roeum, and M.~R. Rosiek, ``New classes of frame wavelets
  for applications,'' in \emph{SPIE's 1995 Symposium on OE/Aerospace Sensing
  and Dual Use Photonics}.\hskip 1em plus 0.5em minus 0.4em\relax International
  Society for Optics and Photonics, 1995, pp. 923--934.

\bibitem{r15}
Q.~Zhang and A.~Benveniste, ``Wavelet networks,'' \emph{Neural Networks, IEEE
  Transactions on}, vol.~3, no.~6, pp. 889--898, 1992.

\bibitem{r16}
G.~Strang and T.~Nguyen, ``Wavelets and filter banks,'' \emph{Wavelets: Theory
  and Applications: Theory and Applications}, p.~38, 1995.

\bibitem{r17}
J.~Catal{\~a}o, H.~Pousinho, and V.~Mendes, ``Hybrid wavelet-pso-anfis approach
  for short-term wind power forecasting in portugal,'' \emph{Sustainable
  Energy, IEEE Transactions on}, vol.~2, no.~1, pp. 50--59, 2011.

\bibitem{r18}
C.~W. Potter and M.~Negnevitsky, ``Very short-term wind forecasting for
  tasmanian power generation,'' \emph{Power Systems, IEEE Transactions on},
  vol.~21, no.~2, pp. 965--972, 2006.

\bibitem{r19}
Y.~V. Makarov, P.~V. Etingov, J.~Ma, Z.~Huang, and K.~Subbarao, ``Incorporating
  uncertainty of wind power generation forecast into power system operation,
  dispatch, and unit commitment procedures,'' \emph{Sustainable Energy, IEEE
  Transactions on}, vol.~2, no.~4, pp. 433--442, 2011.

\bibitem{r20}
C.~W. Potter and M.~Negnevitsky, ``Very short-term wind forecasting for
  tasmanian power generation,'' \emph{Power Systems, IEEE Transactions on},
  vol.~21, no.~2, pp. 965--972, 2006.

\bibitem{r21}
W.~Sweldens, ``The lifting scheme: A construction of second generation
  wavelets,'' \emph{SIAM Journal on Mathematical Analysis}, vol.~29, no.~2, pp.
  511--546, 1998.

\bibitem{r22}
S.~Mallat, \emph{A wavelet tour of signal processing: the sparse way}.\hskip
  1em plus 0.5em minus 0.4em\relax Access Online via Elsevier, 2008.

\bibitem{r23}
A.~Lapedes and R.~Farber, ``A self-optimizing, nonsymmetrical neural net for
  content addressable memory and pattern recognition,'' \emph{Physica D:
  Nonlinear Phenomena}, vol.~22, no.~1, pp. 247--259, 1986.

\bibitem{r24}
J.~T. Connor, R.~D. Martin, and L.~Atlas, ``Recurrent neural networks and
  robust time series prediction,'' \emph{Neural Networks, IEEE Transactions
  on}, vol.~5, no.~2, pp. 240--254, 1994.

\bibitem{r25}
R.~J. Williams and D.~Zipser, ``A learning algorithm for continually running
  fully recurrent neural networks,'' \emph{Neural computation}, vol.~1, no.~2,
  pp. 270--280, 1989.

\bibitem{r26}
D.~Zipser and R.~J. Williams, ``Experimental analysis of the real-time
  recurrent learning algorithm,'' \emph{Connection Science}, vol.~1, no.~1, pp.
  87--111, 1989.

\bibitem{r27}
D.~P. Mandic and J.~Chambers, \emph{Recurrent neural networks for prediction:
  learning algorithms, architectures and stability}.\hskip 1em plus 0.5em minus
  0.4em\relax John Wiley \& Sons, Inc., 2001.

\bibitem{r28}
S.~S. Haykin, S.~S. Haykin, S.~S. Haykin, and S.~S. Haykin, \emph{Neural
  networks and learning machines}.\hskip 1em plus 0.5em minus 0.4em\relax
  Prentice Hall New York, 2009, vol.~3.

\bibitem{r29}
R.~L. Claypoole~Jr, R.~G. Baraniuk, and R.~D. Nowak, ``Adaptive wavelet
  transforms via lifting,'' in \emph{Acoustics, Speech and Signal Processing,
  1998. Proceedings of the 1998 IEEE International Conference on},
  vol.~3.\hskip 1em plus 0.5em minus 0.4em\relax IEEE, 1998, pp. 1513--1516.

\bibitem{r30}
C.~Napoli, F.~Bonanno, and G.~Capizzi, ``An hybrid neuro-wavelet approach for
  long-term prediction of solar wind,'' in \emph{IAU Symposium}, no. 274.\hskip
  1em plus 0.5em minus 0.4em\relax Cambridge Univ Press, 2010, pp. 247--249.

\bibitem{r31}
------, ``Exploiting solar wind time series correlation with magnetospheric
  response by using an hybrid neuro-wavelet approach,'' in \emph{IAU
  Symposium}, no. 274.\hskip 1em plus 0.5em minus 0.4em\relax Cambridge Univ
  Press, 2010, pp. 250--252.

\bibitem{r32}
G.~Capizzi, F.~Bonanno, and C.~Napoli, ``A wavelet based prediction of wind and
  solar energy for long-term simulation of integrated generation systems,'' in
  \emph{Power Electronics Electrical Drives Automation and Motion (SPEEDAM),
  2010 International Symposium on}.\hskip 1em plus 0.5em minus 0.4em\relax
  IEEE, 2010, pp. 586--592.

\bibitem{r33}
G.~Capizzi, C.~Napoli, and L.~Patern{\`o}, ``An innovative hybrid neuro-wavelet
  method for reconstruction of missing data in astronomical photometric
  surveys,'' in \emph{Artificial Intelligence and Soft Computing}.\hskip 1em
  plus 0.5em minus 0.4em\relax Springer, 2012, pp. 21--29.

\bibitem{r34}
M.~M. Gupta, L.~Jin, and N.~Homma, \emph{Static and dynamic neural networks:
  from fundamentals to advanced theory}.\hskip 1em plus 0.5em minus 0.4em\relax
  John Wiley \& Sons, 2004.

\bibitem{r35}
W.~Yan, ``Toward automatic time-series forecasting using neural networks,''
  \emph{Neural Networks and Learning Systems, IEEE Transactions on}, vol.~23,
  no.~7, pp. 1028--1039, 2012.

\bibitem{r36}
M.~Qi and G.~P. Zhang, ``Trend time--series modeling and forecasting with
  neural networks,'' \emph{Neural Networks, IEEE Transactions on}, vol.~19,
  no.~5, pp. 808--816, 2008.

\bibitem{r37}
R.~V. Borges, A.~d'Avila Garcez, and L.~C. Lamb, ``Learning and representing
  temporal knowledge in recurrent networks,'' \emph{Neural Networks, IEEE
  Transactions on}, vol.~22, no.~12, pp. 2409--2421, 2011.

\bibitem{r38}
S.~Yilmaz and Y.~Oysal, ``Fuzzy wavelet neural network models for prediction
  and identification of dynamical systems,'' \emph{Neural Networks, IEEE
  Transactions on}, vol.~21, no.~10, pp. 1599--1609, 2010.

\bibitem{r39}
M.~A. Gonzalez-Olvera and Y.~Tang, ``Black-box identification of a class of
  nonlinear systems by a recurrent neurofuzzy network,'' \emph{Neural Networks,
  IEEE Transactions on}, vol.~21, no.~4, pp. 672--679, 2010.

\bibitem{rr}
G.~Capizzi, C.~Napoli, and F.~Bonanno, ``Innovative second-generation wavelets
  construction with recurrent neural networks for solar radiation
  forecasting,'' \emph{Neural Networks and Learning Systems, IEEE Transactions
  on}, vol.~23, no.~11, pp. 1805--1815, 2012.

\end{thebibliography}
\end{document}